\documentclass[runningheads]{llncs}
\usepackage{amsmath}
\usepackage[T1]{fontenc}
\usepackage{cite}
\usepackage{hyperref}
\usepackage{float}
\usepackage{booktabs}
\usepackage{graphicx,verbatim}
\usepackage{multirow}
\usepackage{marvosym}
\begin{document}

\title{Hierarchical Part-based Generative Model for Realistic 3D Blood Vessel}

\author{
  Siqi Chen\inst{1} \and
  Guoqing Zhang\inst{1, 2} \and
  Jiahao Lai\inst{1} \and
  Bingzhi Shen\inst{3} \and
  Sihong Zhang\inst{1} \and
  Caixia Dong\inst{4} \and
  Xuejin Chen\inst{5} \and
  Yang Li\inst{1}\textsuperscript{\Letter}
}

\authorrunning{S. Chen et al.}

\institute{
  Shenzhen Key Laboratory of Ubiquitous Data Enabling, 
  Tsinghua Shenzhen International Graduate School, Tsinghua University\\
  \email{csq23@mails.tsinghua.edu.cn, yangli@sz.tsinghua.edu.cn}
  \and
  Pengcheng Laboratory
  \and
  School of Life Science and the Key Laboratory of Convergence Medical Engineering System and Healthcare Technology, Ministry of Industry and Information Technology, Beijing Institute of Technology
  \and
  Institute of Medical Artificial Intelligence, 
  the Second Affiliated Hospital of Xi'an Jiaotong University
  \and
   MoE Key Laboratory of Brain-inspired Intelligent Perception and Cognition, University of Science and Technology of China\\
}

\maketitle

\begin{abstract}
Advancements in 3D vision have increased the impact of blood vessel modeling on medical applications. However, accurately representing the complex geometry and topology of blood vessels remains a challenge due to their intricate branching patterns, curvatures, and irregular shapes. In this study, we propose a hierarchical part-based framework for 3D vessel generation that separates the global binary tree-like topology from local geometric details. Our approach proceeds in three stages: (1) \emph{key graph generation} to model the overall hierarchical structure, (2) \emph{vessel segment generation} conditioned on geometric properties, and (3) \emph{hierarchical vessel assembly} by integrating the local segments according to the global key graph. We validate our framework on real-world datasets, demonstrating superior performance over existing methods in modeling complex vascular networks. This work marks the first successful application of a part-based generative approach for 3D vessel modeling, setting a new benchmark for vascular data generation. The code is available at: \href{https://github.com/CybercatChen/PartVessel.git}{https://github.com/CybercatChen/PartVessel.git}.

\end{abstract}

\section{Introduction}

With the rapid progress in computer graphics, 3D visual technology has significantly enhanced the ability to accurately model medical data \cite{survey,3dsurvey1,3dsurvey2}. In particular, 3D vessel generation is crucial for precisely simulating the complex structure of the vasculature, enabling a range of medical applications, from diagnostic assessments \cite{hochmuth2002comparison} to treatment planning \cite{lyu2022reta}. It also supports critical tasks such as preoperative simulations \cite{paetzold2021whole} and medical image analysis \cite{neron}, facilitating more accurate and effective decision-making. Moreover, this technology generates detailed datasets that can be applied to downstream tasks like vessel segmentation \cite{zhang2023topology,dong2023novel} and labeling \cite{labeling}, further improving automated analysis.

\begin{figure}[!t]
    \centering
    \includegraphics[width=0.95\linewidth]{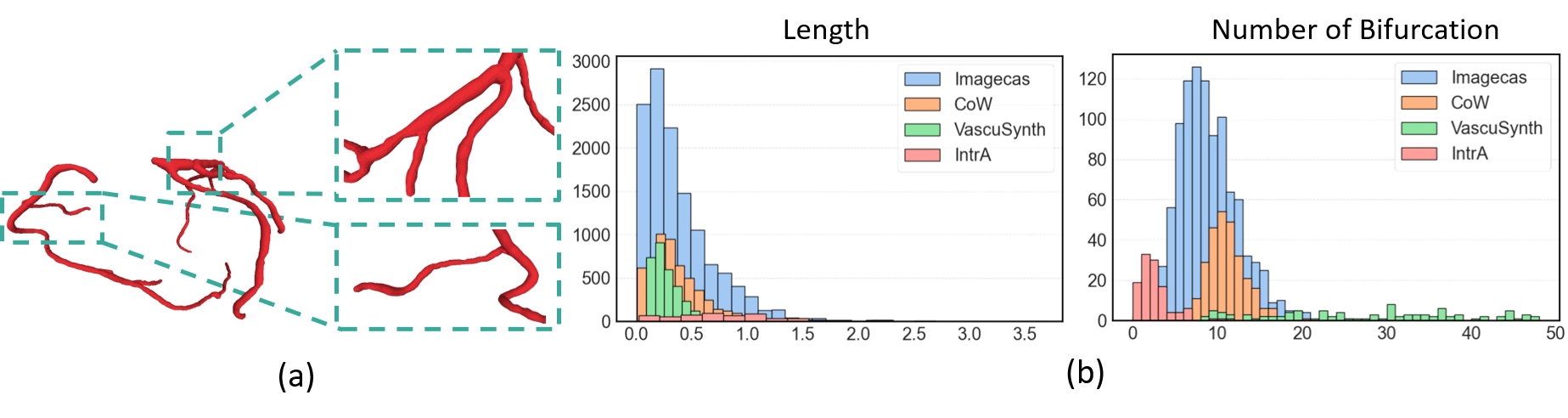}
    \caption{\textbf{(a)} Visualization of a Real-world Coronary Artery Dataset. The vascular network displays a hierarchical, tree-like structure, with localized curvatures and complex branching patterns. \textbf{(b)} The histograms of vessel length and number of bifurcations for four different datasets. 
 }
    \label{fig1}
\end{figure}

Unlike common 3D modeling approaches for objects with regular shapes \cite{shapenet}— such as chairs, tables, or airplanes, which typically feature fixed and predictable structures — vascular modeling poses unique challenges. Those conventional methods \cite{pointdiffusion,lion,pointflow}, often designed for rigid and uniform objects, are not well suited for modeling blood vessels. As illustrated in Figure~\ref{fig1}, the vasculature is characterized by a high degree of complexity, with numerous bifurcations that vary in both number and location. Additionally, the blood vessels exhibit intricate curvature and irregular, non-uniform characteristics. The morphological diversity of vascular structures in real datasets further compounds these challenges. Such complexity demands a model that can effectively capture fine details and accurately represent the diversity of vascular geometries.

To model vascular structures in detail, several methods have been proposed. Point cloud-based approaches designed for common objects \cite{pointdiffusion}, while effective in representing 3D objects, struggle with capturing the geometry of tubular and elongated structures due to their discrete nature. TreeDiffusion \cite{treediffusion} utilizes implicit neural fields for modeling anatomical trees, but its flexibility and accuracy are limited in capturing complex vascular geometries. VesselVAE \cite{vesselvae} employs skeletal graphs to effectively capture vascular structure. However, it generates the entire vascular network without explicitly addressing the unique characteristics of individual branches. As a result, it performs well for simpler vessels with fewer bifurcations, but its fidelity declines for vessels with many branches.

Although these methods have considered the geometric properties of blood vessels, an effective model must capture both the global and local characteristics. Globally, most vessels follow a tree-like hierarchical structure, with endpoints and bifurcations largely defining their organization. Locally, blood vessels exhibit similar geometric shapes, with variations in radius and length, but they can be viewed as segments of tubular curves. Based on these observations, we propose a Hierarchical Part-based Vessel Generative Framework, where the global vascular structure is represented by a tree-shaped key graph based on the vessel skeleton, and local segments are modeled as sequential curves. This decomposition naturally captures both the branching topology and local geometric details of the vasculature. Our contributions are as follows: (1) To the best of our knowledge, this is the first work to employ a part-based approach for 3D vessel generation. (2) We explicitly represent the global structure of vessels as a key graph and local segments as sequential curves, significantly enhancing vascular modeling detail. (3) We validate our approach on real-world datasets, demonstrating superior performance over existing methods in modeling complex vascular networks.

\section{Related Work}
\paragraph{\textbf{Vessel Generation.}}
Although deep learning has been applied to vascular generation, research in this domain remains limited. Wolterink et al. \cite{vesselgan} first employed GANs to generate single-vessel representations sequentially. Feldman et al. \cite{vesselvae} later introduced VesselVAE to model branching vascular structures, but it can only handle a small number of bifurcations. Sinha et al. \cite{treediffusion} explored implicit neural representations (INRs) and diffusion models for vessel generation, yet fidelity remains a challenge. While these methods offer valuable insights, the complexity of the task still limits practice, highlighting the need for further work to enhance both fidelity and structural accuracy.


\paragraph{\textbf{3D Part-based Shape Modeling.}}
Part-based methods in deep learning decompose complex 3D shapes into semantically meaningful components, learned separately and then assembled into a complete structure \cite{chaudhuri2020learning,partvae}. GRASS \cite{grass} and StructureNet \cite{structurenet} first underscored the importance of hierarchical structure, while CompoNet \cite{Componet} increased shape diversity via transformations of parts and their combinations. Drawing on these approaches, we note a parallel to blood vessels, which exhibit a global tree-like structure and locally tubular geometry.


\section{Methodology}
\paragraph{\textbf{Overview.}}

\begin{figure}[!t]
    \centering
    \includegraphics[width=0.95\linewidth]{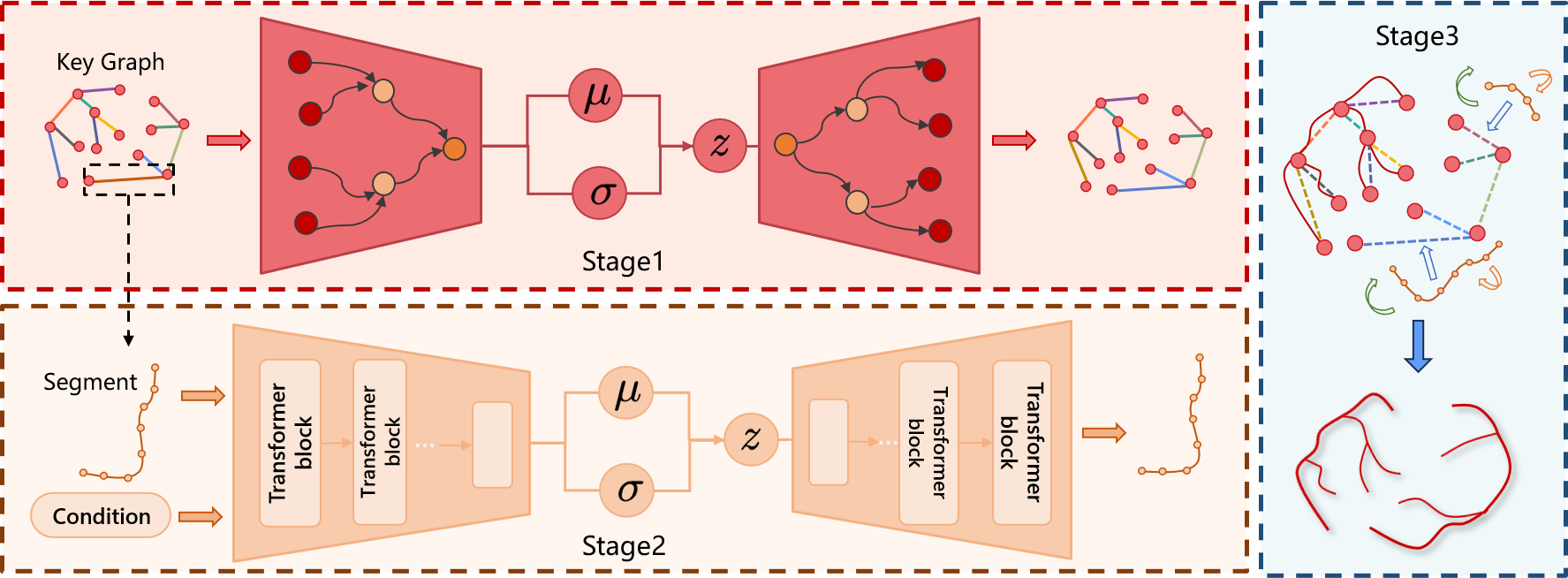}
    \caption{Overall pipeline of our method. \textbf{Stage 1.} Key Graph Generation: learn a global hierarchical tree. \textbf{Stage 2.} Vessel Segment Generation: model local 3D curve based on geometric conditions. \textbf{Stage 3.} Hierarchical Vessel Assembly: reconstruct the vessel skeleton by assembling segments based on the global layout. }
    \label{fig2}
\end{figure}

We adopt a skeleton representation and construct a binary-tree \emph{key graph} based on the skeleton’s bifurcation and terminal points. In this key graph, each edge corresponds to a distinct vessel segment.
Our part-based generative model comprises three stages, as illustrated in Figure~\ref{fig2}. 
The first stage learns the global binary tree structure. The second stage models each vessel segment. Finally, the third stage assembles the synthesized segments according to the key graph to reconstruct the complete vessel. In what follows, we provide a detailed explanation of the methods employed in each stage.

\paragraph{\textbf{Stage 1. Key Graph Generation.}}
Recursive Autoencoders (RAE) were initially introduced by \cite{recursive1,recursive2} and later applied to object modeling by \cite{grass}, as well as vessel modeling by \cite{vesselvae}. To extend RAE into a generative framework, we employ a \emph{Recursive Variational Autoencoder (RVAE)} to model and synthesize a key graph representation of the vascular network. Each node’s attributes consist of three parts: (1) 3D spatial coordinates [$x,y,z$], (2) direction [$n_x,n_y,n_z$] of the local segment, and (3) a geometric descriptor $C=[\ell, \delta, \kappa, \rho]$ that characterizes the local segment's properties, which will be explained in the following stage.

\textbf{Encoding phase.}
Starting from the leaf nodes and moving upward, we aggregate child node features into their parent node. 
Let $v_{parent}$ be a parent node's attribute, and  $h_{left}$ and $h_{right}$ be the hidden states of its left and right children, respectively. 
The parent's hidden state $h_{parent}$ is computed as follows:

\begin{equation}
h_{parent}=\mathrm{MLP} (\mathrm{concat}[v_{parent},h_{left},h_{right}]).
\end{equation}

By repeatedly performing this operation up the tree, we eventually reach the root node $\mathbf z_{root}$, which serves as a global latent embedding for the entire graph.


\textbf{Decoding phase.}
We reverse the procedure to reconstruct node attributes from the root down to the leaves. For each parent node with hidden state $h_{\mathrm{parent}}$, we first use a classifier $\mathbf y = \mathrm{NodeCLS}(\hat{h}_{parent})$ to determine whether left and/or right children exist. If a left child is predicted, its attribute is computed by:
\begin{equation}
    \hat{v}_{left} = \mathrm{MLP}_{\mathrm{left}}(\hat{h}_{parent}).
\end{equation}

We then obtain the left child’s hidden representation by:
\begin{equation}
    \hat{h}_{left} = \mathrm{MLP}(\mathrm{concat}[\hat{h}_{parent}, \hat{v}_{left}]).
\end{equation}

Similarly, if a right child is predicted, we compute $\hat{v}_{right}$ and $\hat{h}_{right}$ in the same manner. This recursive process continues until all nodes are reconstructed.

\textbf{Loss function.}
The total loss includes three terms: 
an \(\mathrm{MSE}(v,\hat{v})\) for node attribute reconstruction,
a \(\mathrm{CrossEntropy}(\hat{\mathbf y}, \mathbf y)\) for node-level classification,
and a KL divergence to regularize the latent space. These terms together define the final objective:
\begin{equation}
    \mathrm{Loss}=\mathrm{MSE} (\hat{v}, v)+
    \mathrm{CrossEntropy} (\hat{\mathbf y}, \mathbf y)+
    D_{\mathrm{KL}} (q({\mathbf z_{root}})\| p(\mathbf z_{root})).
\end{equation}

\begin{figure}[!t]
    \centering
    \includegraphics[width=0.95\linewidth]{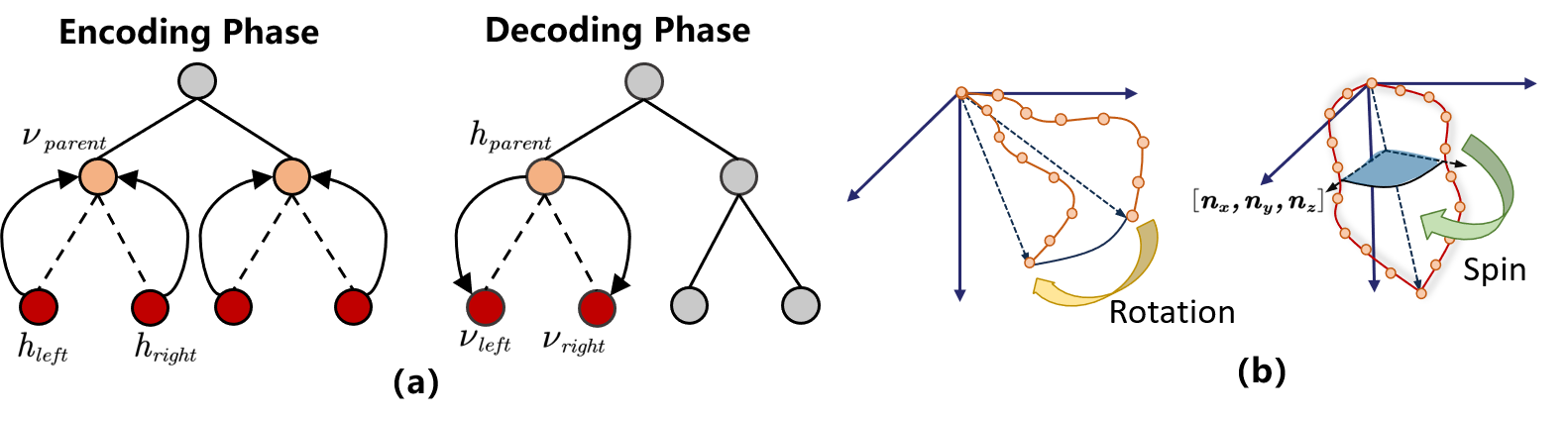}
    \caption{\textbf{(a)} The encoding and decoding process of the model in \emph{Stage 1}. \textbf{(b)} The two types of rotation processes in \emph{Stage 3}.}
    \label{fig3}
\end{figure}
\paragraph{\textbf{Stage 2. Vessel Segment Generation.}}
For each vessel segment identified in the key graph, we represent the skeleton as an ordered sequence in 3D space. Each point along the sequence is described by \(\mathbf x=[x,y,z,r]\), where \(r\) is the vessel radius. To capture the segment's shape, we introduce the geometric descriptor $C=[\ell, \delta, \kappa, \rho]$ as conditional variables, corresponding to key graph attributes. Specifically, \( \ell \) represents the segment's length, \( \delta \) is the straight-line distance between the endpoints, \( \kappa \) quantifies the segment’s curvature, and \( \rho \) indicates the segment’s depth within the binary tree, accounting for variations at different branching levels. These features collectively capture both local geometry and broader structural attributes, enabling accurate generative modeling of vessel segments.

\textbf{Encoding \& decoding.} We adopt a Transformer-based variational framework in which each point is treated as a token. By conditioning on the geometric descriptor $C$, the Transformer can synthesize more realistic vessel segments.
The encoder maps the input sequence to a latent representation, and the decoder then generates new sequences from this latent space.

\textbf{Loss function.}
The total loss includes three terms: an \(\mathrm{MSE}(\mathbf x,\hat{\mathbf x})\) that measures the reconstruction error, a \(\mathrm{CrossEntropy}(\ell,\hat{\ell})\) to ensure the sequence's length, and a KL divergence for generation. Hence, the objective is given by:
\begin{equation}
    \mathrm{Loss}=\mathrm{MSE} (\mathbf x,\hat{\mathbf x})+
    \mathrm{CrossEntropy} (\ell,\hat{\ell})+
    D_{\mathrm{KL}} (q(\mathbf z|\mathbf x,C)\| p(\mathbf z|C)).
\end{equation}

\paragraph{\textbf{Stage 3. Hierarchical Vessel Assembly.}}
In this final stage, we sample a latent vector from the latent space in \emph{Stage 1}, and decode it into a key graph, and assemble this graph with the individually synthesized vessel segments from \emph{Stage 2} to build the complete skeleton in the same frame. Specifically, we employ a depth-first search traversal starting from the root node of the key graph. At each step of the traversal, we first attach the corresponding vessel segment by applying scaling and translation, ensuring spatial alignment and orientation. 

We then rotate the segment so its local direction aligns with the orientation [\(n_x,n_y,n_z\)] derived from the key graph. Figure \ref{fig3}(b) shows the two kinds of rotation processes. These rotations ensure directional consistency with the overall vessel geometry.
By progressively attaching each segment via the above transformations, we obtain a complete vessel skeleton. Finally, following the mesh reconstruction method in \cite{zhang2024geometric} , we leverage the predicted radius of each segment to reconstruct the final 3D vessel surface from the skeleton.

\section{Experiments}
\paragraph{\textbf{Dataset and Data Preparation.}}
We conducted experiments on two real-world datasets and one synthetic dataset, all of which are publicly available. (a) ImageCAS \cite{zeng2023imagecas}: 1,000 real 3D CCTA scans of coronary arteries, presenting significant anatomical variability. We are the first study to address this challenging dataset. (b) VascuSynth \cite{hamarneh2010vascusynth}: A synthetic dataset comprising 120 generated 3D vascular trees with varying numbers of bifurcations. (c) Processed CoW \cite{vesselgraph,cow}: 300 processed 3D vascular meshes of intracranial arteries. All datasets used 90\% for training and 10\% for testing.

We perform a series of preprocessing steps on 3D volumes, to extract skeletons with radius information and derive key graphs. First, we apply morphological operations to the binary label volumes to obtain both the skeleton and its corresponding surface.
Next, we adopt an adaptive mapping \cite{zhang2024geometric} method to construct a key graph and build a maximum spanning tree, where the root, leaf, and branch nodes are identified as vertices of the key graph.

\paragraph{\textbf{Baselines and Metrics.}}
To evaluate our proposed method, we selected three baseline models for comparison. First, we included a state-of-the-art point cloud generation model \cite{pointdiffusion}. Meanwhile, we incorporated two models specifically designed for vessel generation: TreeDiffusion \cite{treediffusion} (D=128, L=5) and VesselVAE \cite{vesselvae}.
To provide a comprehensive evaluation of vessel reconstruction and generation, we utilize metrics based on both point clouds and graph representations. For point clouds sampled from meshes, we report Jensen-Shannon Divergence (JSD) and Chamfer distance (CD) \cite{2018learning,pointflow} to evaluate generation and reconstruction quality respectively. For assembled skeleton graphs, we focus on examining the geometric and topological properties of skeleton graphs, including the maximum mean discrepancy for degree distributions (Deg.) and the Laplacian spectrums (Spec.). To accurately assess the reconstruction performance of skeleton graphs, which are essentially 3D geometric graphs, we adopt the Graph Wasserstein distance (GWD) from \cite{streetmover}.

\begin{table}[!b]
\centering
\footnotesize
\renewcommand{\arraystretch}{1.1}
\setlength{\tabcolsep}{6pt} 
\caption{Comparison of vessel reconstruction and generation on point- and graph-based evaluation metrics. CD and JSD are multiplied by $10^3$. The best results are denoted in \textbf{Bold}. The second-best result is \underline{Underlined}.}
\label{table2}
\begin{tabular} {llccccc}
\toprule

\multirow{2}{*}{{\textbf{Dataset}}} & \multirow{2}{*}{{\textbf{Method}}} & \multicolumn{2}{c}{{Point-based}} & \multicolumn{3}{c}{{Graph-based}}\\
\cmidrule(lr){3-4} \cmidrule(lr){5-7}
& & JSD & CD & Deg. & Spec. & GWD \\ 
\midrule

\multirow{4}{*}{ImageCAS} & VesselVAE & 92.0  & 89.8   &  1.259  & 1.165  &  0.141  \\
                        & PointDiffusion & 82.2  & \textbf{1.1} &-&-&- \\
                          & TreeDiffusion &  \textbf{31.9}  & 36.2 &  \underline{1.092}  &  \underline{0.188}   &  \underline{0.087}  \\
                          & Our   &  \underline{50.1}
 &    \underline{24.4}    &  \textbf{0.601} &  \textbf{0.079}  &  \textbf{0.029} \\
\midrule
\multirow{4}{*}{VascuSynth}& VesselVAE & 87.4  & 59.1   &  1.911   &  1.207   &  0.115  \\
                           & PointDiffusion & 87.1  & \textbf{7.0}&-&-&- \\
                           & TreeDiffusion & \underline{42.2}  & 35.1&  \underline{0.445}   &  \textbf{0.099}   &  \underline{0.083}  \\
                           & Our       & \textbf{40.1}  &  \underline{34.8}    &  \textbf{0.190}  &  \underline{0.159}   &  \textbf{0.049}  \\
\midrule
\multirow{4}{*}{CoW}       & VesselVAE   & 88.9  & 35.6  &  1.740   &  1.037   &  \underline{0.025}  \\
                           & PointDiffusion & 91.0  & \textbf{1.0}&-&-&- \\
                           & TreeDiffusion & \underline{47.7}  & 20.9 & \textbf{0.726}   &  \underline{0.277}   &  0.038  \\
                           & Our      & \textbf{44.1}  & \underline{15.1}      &  \underline{1.475}  & \textbf{0.182}   &  \textbf{0.017}  \\
\bottomrule
\end{tabular}
\end{table}

\paragraph{\textbf{Implementation Details.}}

All experiments were conducted using PyTorch on an NVIDIA A800 GPU with the Adam optimizer. The training process is conducted in two sequential stages. In \textit{stage 1}, we set the latent space to 512 dimensions. We started with a learning rate of 0.001, which was reduced by 0.2 every 100 epochs. The model was trained for 20k epochs, with a batch size of 128, converging in approximately 12 hours. In \textit{stage 2}, we set the latent space to 64 dimensions and the batch size to 512. The learning rate was set to 0.0002, training for 2k epochs, which required about 3 hours to train. We set transformer blocks to 4 layers with 4 multi-head self-attention in both the encoder and decoder, and the maximum sequence length was set to 200.

\paragraph{\textbf{Experimental Result.}}
Table \ref{table2} compares our method with three state-of-the-art approaches under point cloud-based and graph-based metrics. Our model achieves competitive performance across most tasks, particularly in graph-based evaluation of structural fidelity and topological consistency. In contrast, VesselVAE struggles with complex branching involving numerous sequential points, resulting in consistently sub-optimal performance. PointDiffusion demonstrates the strongest reconstruction metrics across all datasets. However, it falls short on generation metrics and sample quality, indicating an inability to adequately capture vascular geometry and topology. We also observed that TreeDiffusion reports a high JSD score on the ImageCAS dataset but shows poor qualitative generation results. Because JSD is primarily designed to evaluate the spatial distribution of point clouds, however, it fails to account for the accuracy of their topological structures.

\begin{figure}[!t]
    \centering
    \includegraphics[width=0.95\linewidth]{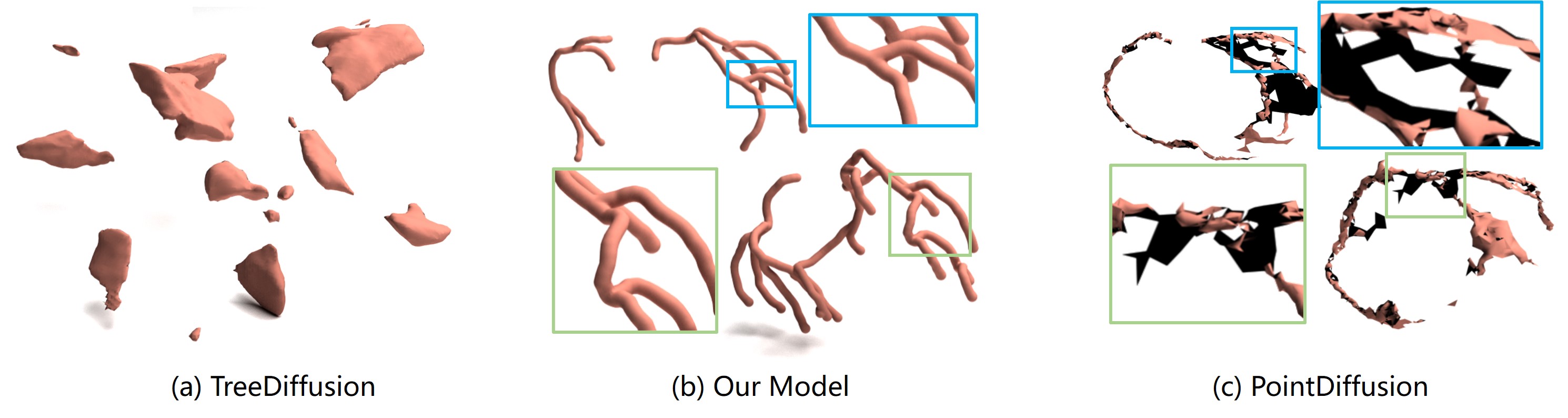}
    \caption{Reconstruction result from three different methods on ImageCAS dataset. Our approach produces more robust and anatomically consistent results compared to point cloud-based and INR-based methods.}
    \label{fig4}
\end{figure}

\begin{figure}[H]
    \centering
    \includegraphics[width=0.95\linewidth]{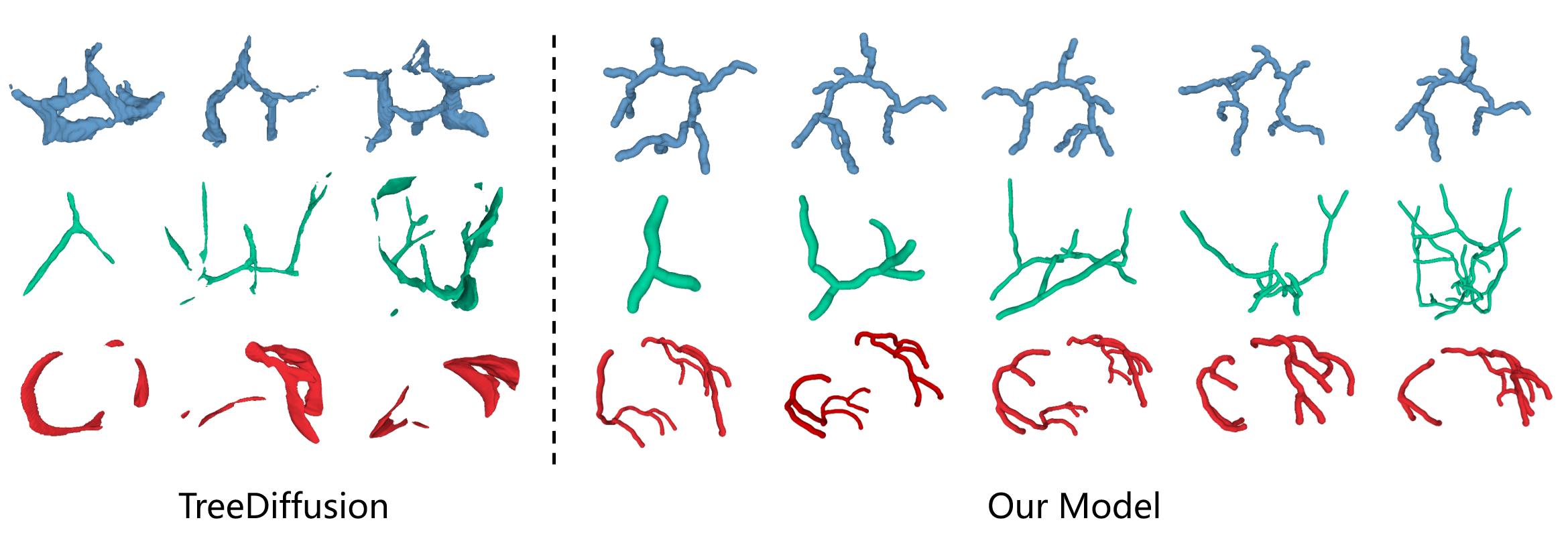}
    \caption{Examples of generation results from TreeDiffusion and our model on CoW, VascuSynth, and ImageCAS datasets (from top to bottom).}
    \label{fig5}
\end{figure}

Although PointDiffusion and TreeDiffusion outperform our method on certain metrics, a visual comparison of their reconstruction and generation samples reveals some fundamental problems in their modeling techniques. Some reconstruction and generation samples are visualized in Figures \ref{fig4} and \ref{fig5}. Particularly, due to the complex data distribution in the ImageCAS dataset, all comparison models failed to successfully generate the basic morphology of the vessels. As shown in Figure \ref{fig4}, our model not only replicates complex branching structures with high accuracy but also captures subtle morphological variations in the vessels.
Point cloud-based methods, by contrast, have difficulty distinguishing between the interior and exterior surfaces of the vessels, leading to multiple holes in the reconstructed meshes and negatively impacting downstream tasks such as vascular analysis. Our skeleton graph-based approach effectively handles complex branching structures and accurately captures morphological variations, underscoring its robustness under real-world conditions. By directly leveraging skeleton and radius information, our method naturally avoids the pitfalls of point cloud-based reconstructions and produces more robust mesh reconstructions for tubular structures.
Figure~\ref{fig5} compares the generative performance of our approach against the highly competitive TreeDiffusion, using TreeDiffusion's best-performing samples. As shown, TreeDiffusion often produces irregular, block-like shapes and disconnected components across all datasets, indicating structural anomalies. In contrast, our model maintains vascular continuity and produces more realistic, anatomically consistent vessel networks.

\section{Conclusion}
In this paper, we propose a hierarchical part-based framework for 3D vessel generation that separates the global tree-like structure from local geometries. Our approach proceeds in three stages: first, we employ a recursive variational autoencoder to construct a key graph to capture the vascular hierarchy. Subsequently, we introduce a transformer-based VAE to synthesize detailed vessel segments, which are then assembled into a complete vessel in the final stage. Experimental results on three public datasets demonstrate that our model consistently preserves vascular continuity and authentic local curve characteristics.

\begin{credits}
\subsubsection{\ackname} This work is supported in part by the Natural Science Foundation of China (Grant 62371270).

\subsubsection{\discintname}
The authors have no competing interests to declare that are relevant to the content of this article.
\end{credits}

\bibliographystyle{splncs04}
\bibliography{ref}

\end{document}